\documentclass[aps,reprint,floatfix,amsmath,amssymb,superscriptaddress]{revtex4-1}
\usepackage{graphicx}
\usepackage{bm}
\def\coloneqq{\mathrel{\mathop:}=}
\newcommand{\argmin}{\mathop{\rm arg~min}\limits}

\begin{document}
%%% Title of paper
\title{Image Analysis Based on Nonnegative/Binary Matrix Factorization}

%%% Authors
%\author{Hinako Asaoka\thanks{asaoka.hinako@is.ocha.ac.jp} and Kazue Kudo\thanks{kudo@is.ocha.ac.jp}}
\author{Hinako Asaoka}
\email[]{asaoka.hinako@is.ocha.ac.jp}
\author{Kazue Kudo}
\email[]{kudo@is.ocha.ac.jp}
\affiliation{Department of Computer Science, Ochanomizu University, Tokyo 112-8610, Japan}

%%% Abstract
\begin{abstract} 
Using nonnegative/binary matrix factorization (NBMF), a matrix can be decomposed into a nonnegative matrix and a binary matrix.
Our analysis of facial images, based on NBMF and using the Fujitsu Digital Annealer, leads to successful image reconstruction and image classification.
The NBMF algorithm converges in fewer iterations than those required for the convergence of nonnegative matrix factorization (NMF), although both techniques perform comparably in image classification.
\end{abstract} 

\maketitle

%%% Ising machines
Solving combinatorial optimization problems is computationally intensive, especially when the size of the problem is enormous.
In recent years, special-purpose hardware devices for solving optimization problems have been developed~\cite{Johnson2011, Gibney2017, Inagaki2016, Yamaoka2016, Aramon2019, Goto2019}.
For example, a D-Wave quantum annealer is a quantum device realizing quantum annealing (QA)~\cite{Johnson2011, Gibney2017}.
QA is known as an algorithm based on quantum mechanics for solving combinatorial optimization problems~\cite{Kadowaki1998}.
Several other Ising machines, which are hardware implementing the Ising model, have also been developed~\cite{Inagaki2016, Yamaoka2016, Aramon2019, Goto2019}.
Ising machines, including quantum annealers, require the formulation of the problems to be solved in the form of the Ising model or quadratic unconstrained binary optimization (QUBO).
Despite this requirement, Ising machines are applied to a wide variety of optimization problems,
which are not limited to typical combinatorial optimization.
Optimization problems also appear in machine learning problems, and
several works have reported the application of Ising machines to machine learning problems~\cite{Neven2012, Benedetti2017, Neukart2018, Li2018, OMalley2018}.

%%% NMBF
Matrix factorization is a method that extracts features from datasets in machine learning analyses.
When matrix factorization is applied to facial images, each face can be represented as a linear combination of basis images that correspond to the facial features.
In the case of nonnegative matrix factorization (NMF), the basis images correspond to facial parts, such as mouths, noses, and eyes~\cite{Lee1999}.
NMF decomposes a matrix into two nonnegative matrices.
Using a similar method called nonnegative/binary matrix factorization (NBMF)~\cite{OMalley2018}, a matrix is decomposed into a nonnegative matrix and a binary matrix.
The nonnegative matrix corresponds to the basis images while the binary matrix represents a combination of the basis images.
Since NBMF is a type of combinatorial optimization problem, Ising machines are useful in solving these problems.

%%% aim of this paper
In this study, we analyze facial images based on NBMF, using the Fujitsu Digital Annealer~\cite{Aramon2019}.
The Digital Annealer is an Ising machine that implements simulated annealing~\cite{Kirkpatrick1983} and exchange Monte-Carlo simulations~\cite{Hukushima1996} to solve fully-connected QUBO problems.
We demonstrate the successful reconstruction of facial images from matrices decomposed by NBMF.
We also perform image classification using NBMF.
After unsupervised machine learning based on NBMF, a binary vector represents the corresponding facial image.
We classify facial images using the distance between them.
Since the distance between binary vectors can be computed more easily than that for real vectors, NBMF is more efficient than NMF in image classification.
Owing to the discreteness of binary vectors, NBMF is expected to prevent overfitting and be advantageous when the number of training data is small~\cite{Li2018}.
Moreover, the utilization of an Ising machine such as the Digital Annealer, accelerates computation using NBMF.

%%% NBMF algorithm
We employ the NBMF algorithm introduced in Ref.~\cite{OMalley2018}.
We prepare an $n \times m$ data matrix $V$ from $m$ images,
where each column of $V$ corresponds to an $n$-pixel gray-scale image.
The elements of $V$ are normalized as $0\le V_{ij} \le 1$.
The aim of NBMF is to find an $n \times k$ nonnegative matrix $W$ and a $k \times m$ binary matrix $H$ such that
\begin{equation}
  V \approx W H,
  \label{eq:v}
\end{equation}
where $W_{ij} \ge 0$ and $H_{ij} \in \{ 0,1 \}$.
The columns of $W$ correspond to the basis images, which are the features of the facial images.
The number of basis images, i.e., facial features, is $k$.
Each column of $H$ represents the combination of basis images which make up the corresponding image in $V$.
Since each column of $H$ is a $k$-dimensional binary vector, it contains considerably smaller amount of data than $V$.

We update $W$ and $H$ alternately, applying the least-squares method.
The initial values of $W_{ij} \ge 0$ and $H_{ij} \in \{ 0,1 \}$ are assigned randomly.
Using the projected gradient method~\cite{Lin2007}, we update $W$ as
\begin{equation}
  W \coloneqq \argmin_{X \in \mathbb{R}^{+n \times k}}
\| V - XH \|_F + \alpha  \| X \|_F,
    \label{eq:w}
\end{equation}
where $\| \cdot \|_F$ and $\alpha$ are the Frobenius distance and a positive constant, respectively.
$H$ is updated by computing
\begin{equation}
  H \coloneqq \argmin_{X \in { \{ 0,1 \} }^{k \times m}}
                    \| V - WX \|_F,
    \label{eq:h}
\end{equation}
using an Ising machine.
Since $H$ is a binary matrix, Ising machines can solve the problem efficiently.
In this work, we use the Digital Annealer,
which accepts only the QUBO formulation. Hence, we transform Eq.~\eqref{eq:h} into the form
\begin{equation}
  f(\bm{q}) = \sum_i a_i q_i + \sum_{i<j} b_{ij} q_{i} q_{j},
  \label{eq:f}
\end{equation}
where $\bm{q} \in { \{ 0,1 \} }^{k}$.
The coefficients $a_i$ and $b_{ij}$ for each column of $V$ are provided as inputs, and the Digital Annealer returns the $\bm{q}$ that minimizes the objective function $f(\bm{q})$.
For the $l$th column of $V$,
\begin{eqnarray}
  a_i &=& \sum_r W_{ri}(W_{ri} - 2 V_{rl}),
  \label{eq:a_i}\\
b_{ij} &=& 2 \sum_r W_{ri} W_{rj}.
  \label{eq:b_ij}
\end{eqnarray}
We repeat the above updating procedure until $W$ converges.
When the Frobenius distance between the updated and previous values of $W$ is less than $10^{-4}$, $W$ is assumed to have converged.

%%% NMF algorithm
We also examine NMF for comparison.
In this case, $H$ is a nonnegative matrix but not a binary one:
$W_{ij} \ge 0$ and $H_{ij} \ge 0$.
We update $W$ and $H$ alternately as follows.
\begin{eqnarray}
  W_{ij} &\leftarrow& W_{ij} \sum_{r} \frac{V_{ir}}{(WH)_{ir}} H_{jr},
  \quad
  W_{ij} \leftarrow \frac{W_{ij}}{\sum_{r} W_{rj}},
    \label{eq:w_nmf}\\
    H_{ij} &\leftarrow& H_{ij} \sum_{r} W_{ri} \frac{V_{rj}}{(WH)_{rj}}.
      \label{eq:h_nmf}
\end{eqnarray}
We repeat the iteration until the Frobenius distance between the updated $W$ and the previous one falls below $10^{-4}$ and convergence is achieved.

%%% numerical conditions
The numerical experiments described below utilize facial images from the Olivetti faces dataset (AT \& T Laboratories Cambridge).
The number of pixels in each image is $n=1024$.
The number of features and the regularization parameter are assigned the values $k=60$
and $\alpha=10^{-6}$, respectively.
Parameter dependence is insignificant. 
When $40\lesssim k \lesssim 80$ and $\alpha \lesssim 10^{-3}$, the behavior observed is similar to the following results. 

\begin{table}[tb]
  \caption{The final average RMSE and the number of iterations.}
  \label{tab:RMSE}
  \begin{tabular}{|c||c|c|c|} \hline
     & & \multicolumn{2}{|c|}{number of iterations} \\ \cline{3-4}
     & RMSE & average & variance \\ \hline \hline
    NBMF & 0.043326 & 35 & 209.6 \\ \hline
    NMF & 0.030264 & 788.8 & 8954.96 \\ \hline
 \end{tabular}
\end{table}

%%% image reconstruction
First, we examine image reconstruction for comparing the performances of NBMF and NMF.
Since we use a dataset consisting of the facial images of 20 people
and the data matrix $V$ includes five different images per person,
 the number of images in $V$ is $m=100$.
The data matrix is approximated as $V\approx WH$, where $H$ is binary for NBMF and nonnegative for NMF.
We calculate the root mean squared error (RMSE) between each image and the corresponding reconstruction and average over 100 images.
In choosing 5 out of 10 facial images randomly per person for constructing $V$, we take an average of 5 trials with different combinations of images.
The final average RMSEs and the number of iterations for updating $W$ and $H$ are summarized in Table~\ref{tab:RMSE}.
Since the RMSE of NMF is smaller than that of NBMF, it produces better  facial image reconstructions than NBMF.
However, the number of iterations required by NBMF is 20 times smaller,  which indicates that the NBMF algorithm converges much faster than NMF.

\begin{figure}[tb]
\centering
\includegraphics[width=6cm]{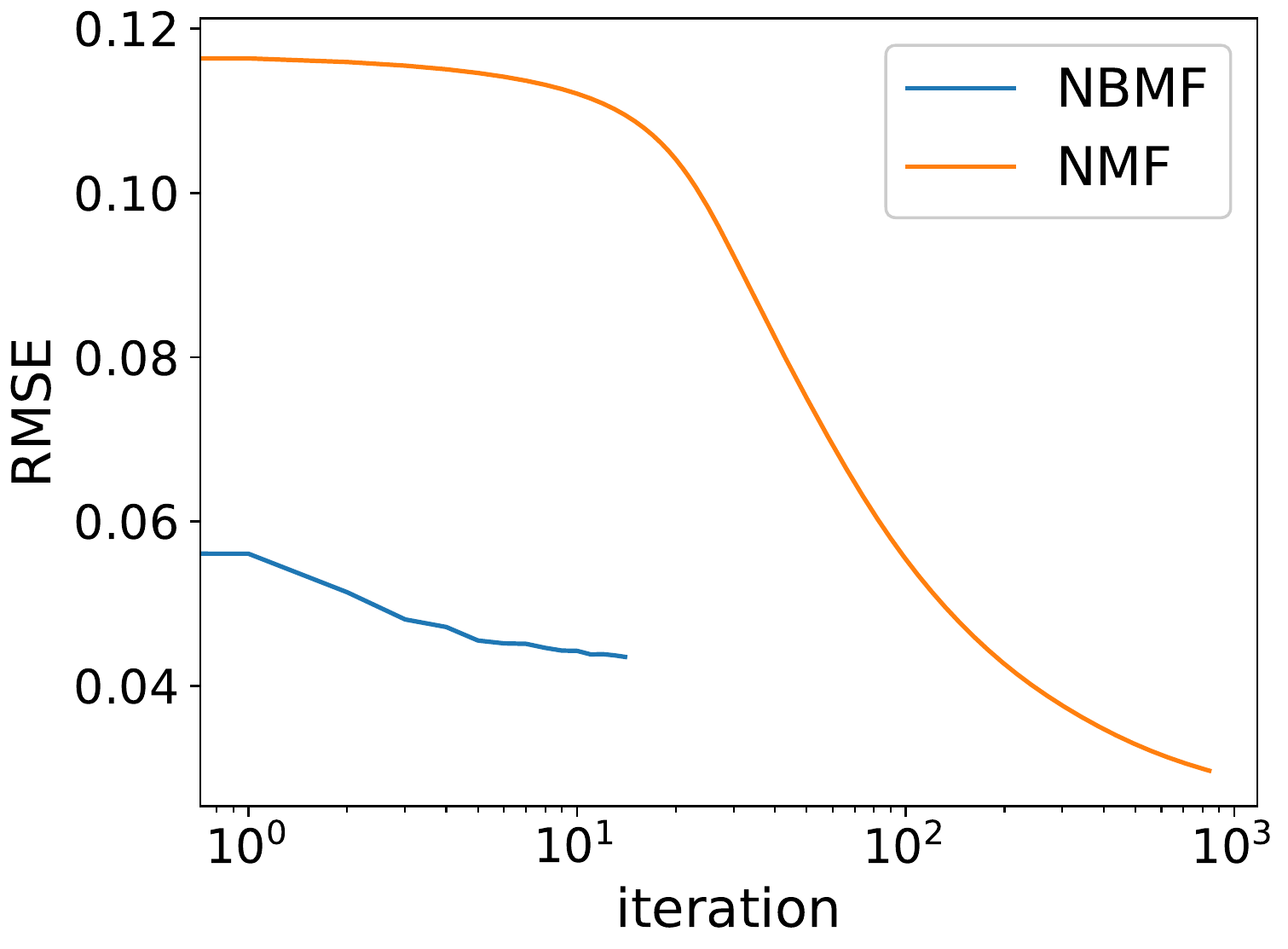}
\caption{(Color online) RMSE between the original and the reconstruction images at each update.
}
\label{fig:g_RMSE}
\end{figure}

Figure~\ref{fig:g_RMSE} plots the RMSEs at each update.
Each curve represents the result corresponding to the smallest iteration number among the five trials.
When the NBMF algorithm is applied, the RMSE is considerably smaller than that of NMF throughout.
In contrast with the gradual change in matrix elements in NMF, sudden changes can occur in NBMF owing to the discreteness of the binary matrix $H$.
The NBMF algorithm converges with a small number of iterations before the RMSE decays sufficiently.
 As a result, the reconstructed image is lower in quality, although the computational costs are reduced.

%%% image classification
Next, we classify facial images using NBMF.
The training data $V$ consists of the facial images of 20 people.
 Five different images per person are included in $V$.
Twenty images that are not included in the training data are used as test data.
However, each test image corresponds to one of the 20 people in the training data.
The matrices $W$ and $H$ obtained by NBMF are used to identify the face each test image corresponds to.
The vector corresponding to a test image is approximated as $\bm{v}\approx W\bm{h}$, where $\bm{h}$ is a binary vector
that indicates the combination of features reproducing the original image.
If $H$ has a column vector similar to $\bm{h}$, the corresponding image and the test image have similar features.
We use the $k$-nearest neighbor algorithm to select the column vectors from $H$ that are similar to $\bm{h}$.
We use the Euclidean distance to measure the similarity.
We calculate the distance between $\bm{h}$ and each column vector of $H$ and choose three nearest neighbors amongst them.
Each column vector has the same label as the corresponding image in $V$.
The label of $\bm{v}$ is the most common among the three ones.

In the numerical experiments using NBMF, 15 out of 20 images were classified correctly, which corresponds to an accuracy of 75 percent.
In the case of NMF, 14 out of 20 images were classified correctly.
Thus, NBMF and NMF are comparable in performance, although NBMF yields lower-quality image reconstruction than NMF.

%%% conclusion
We have demonstrated a successful and efficient NBMF-based image reconstruction and image classification.
The NBMF algorithm converges in fewer iterations than those required for the convergence of the NMF algorithm.
The classification of images using NBMF requires only binary vectors that represent the corresponding images.
The NBMF-based image classification method can be applied to other machine learning problems using extracted features.

\begin{acknowledgments}
HA thanks the METI and IPA for their support through the MITOU Target program.
This work was partially supported by the JSPS KAKENHI Grant Number JP18K11333.
\end{acknowledgments}

%%% Create the reference section using BibTeX:
%\bibliography{nbmf.bib}

\end{document}